\documentclass[conference]{IEEEtran}
\IEEEoverridecommandlockouts
\usepackage{cite}
\usepackage{amsmath,amssymb,amsfonts}
\usepackage{algorithmic}
\usepackage{graphicx}
\usepackage{textcomp}
\usepackage{xcolor}
\usepackage{subcaption}
\def\BibTeX{{\rm B\kern-.05em{\sc i\kern-.025em b}\kern-.08em
    T\kern-.1667em\lower.7ex\hbox{E}\kern-.125emX}}
\usepackage[T1]{fontenc}
\begin{document}

\title{Hardware-in-the-loop simulation of a UAV autonomous landing algorithm implemented in SoC FPGA\\
\thanks{The work presented in this paper was supported by the AGH University of Science and Technology project no. 16.16.120.773}
}


\author{\IEEEauthorblockN{Hubert Szolc}
\IEEEauthorblockA{\textit{Embedded Vision Systems Group} \\ \textit{Computer Vision Laboratory}\\ \textit{Department of Automatic Control and Robotics} \\
\textit{AGH University of Science and Technology}\\
Krakow, Poland \\
szolc@agh.edu.pl} 
\and
\IEEEauthorblockN{Tomasz Kryjak, Senior Member, IEEE}
\IEEEauthorblockA{\textit{Embedded Vision Systems Group} \\ \textit{Computer Vision Laboratory}\\ \textit{Department of Automatic Control and Robotics} \\
\textit{AGH University of Science and Technology}\\
Krakow, Poland \\
tomasz.kryjak@agh.edu.pl} 
}

\maketitle

\begin{abstract}
This paper presents a system for hardware-in-the-loop (HiL) simulation of unmanned aerial vehicle (UAV) control algorithms implemented on a heterogeneous SoC FPGA computing platforms.
The AirSim simulator running on a PC and an Arty Z7 development board with a Zynq SoC chip from AMD Xilinx were used.
Communication was carried out via a serial USB link.
An application for autonomous landing on a specially marked landing strip was selected as a case study.
A landing site detection algorithm was implemented on the Zynq SoC platform.
This allowed processing a $1280 \times 720$ @ 60 fps video stream in real time.
Performed tests showed that the system works correctly and there are no delays that could negatively affect the stability of the control.
The proposed concept is characterised by relative simplicity and low implementation cost. 
At the same time, it can be applied to test various types of high-level perception and control algorithms for UAV implemented on embedded platforms.
We provide the code developed on GitHub, which includes both Python scripts running on the PC and C code running on Arty Z7.

\end{abstract}


\section{Introduction}

In recent years, there has been a significant increase in the popularity of unmanned aerial vehicles (UAVs), commonly referred to as ''drones''. 
This is due to the extremely broad horizon of their potential applications (e.g. military operations, transport of goods, inspection of transmission networks -- e.g. gas pipeline).
Most often, an operator is responsible for the execution of individual missions, who -- while remaining on the ground -- uses a wireless link, including image transmission from the camera mounted on the drone, to control the vehicle. 
However, this is a significant limitation of the drone's functionality. 
A positive result of the mission depends mainly on the skills of a given pilot, and improving them requires additional training costs. 
Furthermore, the flight range is strictly limited by the quality of the video stream transmission, and if it is disturbed, the vehicle practically loses all its operational capacity. 
This poses a risk of serious damage or even total destruction. 
In view of this, many research centres around the world (both for university, commercial, and military purposes) are currently working intensively on the automation of drone flight so that they can perform their missions autonomously \cite{Elmokadem2021}.


The replacement of the operator by an automatic control system is generally done in one of two ways.
In the first approach, control signal values are calculated by an algorithm running on the so-called ''ground station'' (e.g. PC class computer) and then transmitted to the flying unit.
However, this method does not solve the issue of limitation resulting from the transmission range, but at the same time, there are no significant restrictions on the available computing power.
In view of this, efforts are also being made to implement control algorithms on embedded platforms that can be integrated directly into the drone.
As a result, the vehicle can explore a much larger area, and the only limitation is a finite energy resource in the form of a battery with a specific capacity.
It should be noted that in this scenario the available computing power is significantly limited; hence, the need for careful algorithm selection and design, as well as the use of an appropriate computing platform.


In this context, it is worth mentioning the heterogeneous SoC (System-on-Chip) platforms, which consist of different computing units, such as CPUs, GPUs (Graphical Processing Units), and FPGAs (Field Programmable Gate Array).
As a result of their integration, they provide high computing performance, due to the possibility of applying domain-specific computations, with relatively low power consumption.
In particular, they enable parallel handling of different data streams (e.g. from camera, IMU (Inertial Measurement Unit) and LiDAR).
With this in mind, it can be concluded that SoCs are very good embedded platforms for implementing control algorithms for UAVs.


However, using SoCs platforms for implementing control algorithms for UAVs, requires the development of appropriate test tools.
They should best mimic the target conditions of drone operation, which is usually a dynamically changing 3D environment.
However, at the same time, these tools must guarantee the highest level of safety, both for the environment and for the vehicle itself.
One of the popular solutions to this problem is the Hardware-in-the-loop (HiL) simulation.
In HiL, the control algorithm runs on the target embedded platform, placed in a specific test harness.
It usually consists of a virtual 3D graphical environment and algorithms responsible for simulating various sensors that are part of the target vehicle's equipment (e.g. camera, IMU, LiDAR).
In this way, it is possible to check the operation of the designed perception and control algorithms, on the target computing platform, without risking damage to the drone.


In our work, we propose a HiL simulation for the case of using a heterogeneous SoC FPGA computing chip as an embedded platform for a drone control algorithm.
Our solution is mainly characterised by its efficiency, but also by its relative simplicity of the setup.
Its implementation requires only a~PC-class computer (preferably with a powerful GPU) as the host for a simulation environment, a multimedia interface (e.g. HDMI) to transmit the video stream and a simple serial link (e.g. USB) to exchange the remaining data (control signals).
This makes our proposed HiL simulation applicable in almost any situation of using an SoC FPGA chip to control a drone.


The main contribution of this paper is the concept and implementation of an easily achievable HiL simulation for heterogeneous SoC FPGA systems.
To the best of our knowledge, similar attempts have not yet been widely discussed in the scientific literature.
Our results demonstrate that this is a valuable approach that is worth using in the design of analogous control systems.
Therefore, we make the codes we used, together with a detailed description of how they can be used, available in a GitHub repository\footnote{https://github.com/vision-agh/HiL\_drone\_landing}.


The reminder of this paper is organised as follows.
Section \ref{sec:prev_work} presents previous work on HiL simulation for unmanned aerial vehicles.
In Section \ref{sec:proposed_system} we describe the structure of our system and how it works.
Its evaluation is in turn presented in Section \ref{sec:evaluation}.
The last section contains our conclusions and further work to be carried out using the proposed system.


\section{Related work}
\label{sec:prev_work}

The use of HiL simulations to verify UAV control algorithms has been addressed in several research paper.
However, most of them focus on low-level controller testing (e.g. Pixhawk).


An example of this type of research is the article \cite{gade2016}.
The authors propose a HiL simulation for the verification of an obstacle avoidance controller intended for Miniature Aerial Vehicles (MAVs).
A programme simulating the drone and its environment is prepared using the MATLAB-Simulink package and run on a real-time computer.
It provides the necessary data to the Pixhawk controler under test.
The control signals it produces are fed back to the real-time computer via actuators (motors without propellers), thus closing the test loop.
The authors used the prepared simulation to test the obstacle avoidance algorithm in an urban environment.
Based on the results obtained in this way, the authors conclude the correct operation of the proposed control algorithm, as well as the HiL simulation itself, whose results were consistent with those obtained in the ''off-line'' tests.


Another example of this type of research is the work of \cite{gong2019}.
The authors consider an extended HiL simulation.
It includes a~fault model that generates disturbances at the input and output of the drone dynamics model.
This is intended to increase the realism of the generated data.
As in the previous example, the authors use a special computer (NI PXIe-8880) to run the simulation.
In this case, it is responsible for modelling the actuators, drone dynamics, and sensors.
The whole simulation is supervised by a computer, which also acts as a server.
The Pixhawk controller is tested again.
According to the authors, the results obtained are consistent with the results of experiments performed directly in Simulink, which confirms the effectiveness of the proposed solution.


The problem of using HiL simulation to verify a higher-level control algorithm is, in turn, considered in the work \cite{Kumar2010}.
The authors propose a~landing algorithm based on tracking reference trajectories generated using the near-minimum-time manoeuvre method.
They tested the developed solution using the FlightGear simulator, which generated a~virtual 3D graphical environment.
The control algorithm was implemented on a~PC104 board with a~QNX Neutrino RTOS (Real Time Operating System).
The authors used a~special HiL bridge module to communicate with the simulator.
The tests carried out demonstrated the correct operation of the proposed algorithm.


A slightly more advanced approach to the topic discussed is presented in \cite{dai2021}.
The authors propose a comprehensive concept for the standardisation of simulation experiments.
Its first element would be a unified framework for modelling different types of UAVs (including their components).
According to the authors, this would facilitate the conduct of HiL simulations.
For this purpose, they propose to use, among other things, parallel computing in an FPGA.
These enable the simulation of multiple sensors simultaneously (e.g. GPS, barometer, or accelerometer).
The authors also propose to simulate the environment in which the vehicle is flying.
They are developing their own solution for this purpose, based on Unreal Engine 4 (the same as used by the AirSim simulator).
The proposed system has been subjected to comparative tests, in which the simulated data were compared with those obtained from real flights.
According to the authors, the results obtained prove the correct operation of their solution.


The use of FPGAs for HiL simulations has also been considered in the paper \cite{wang2017}.
The authors compare the altitude estimates obtained using the direction cosine matrix (DCM) algorithm. 
It was first implemented in the MATLAB-Simulink package and then generated with HDL Coder to an FPGA chip.
Readings from a real drone were used as reference for both cases.
It was suspended on a rope using two Eye Fish joints in a way that guaranteed freedom of movement.
On the basis of the results obtained, the authors conclude that the latency of the simulation carried out using the FPGA chip is significantly lower than that of the MATLAB-Simulink environment.


The described work indicates that HiL simulation is an important issue for the evaluation of UAV control algorithms.
In many cases, the authors use advanced and at the same time expensive technologies for its implementation, such as highly efficient computers with real-time operating systems or special cards for data acquisition.
There is also a noticeable interest in FPGAs as elements of the test harness.
In this context, the possibilities of parallel computing are exploited, which, among other things, facilitate the simultaneous simulation of different sensors.

Furthermore, analysis of the scientific literature indicates that there are no solutions in which an attempt has been made to significantly minimise the cost of implementing the entire HiL simulation.
Also, no research has yet been carried out on the development of such a system in the case of a~control algorithm implemented on a heterogeneous SoC FPGA platform



\section{The proposed system}
\label{sec:proposed_system}

\begin{figure}[!t]
    \centerline{\includegraphics[width=0.45\textwidth]{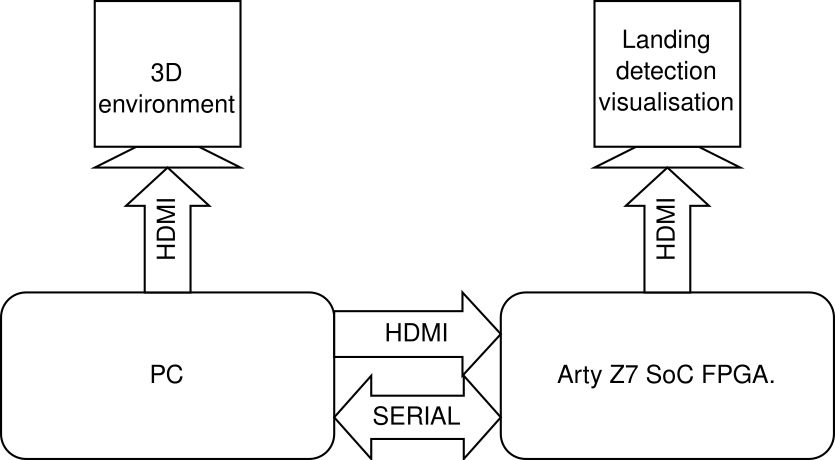}}
    \caption{Scheme of the proposed system. The AirSim simulator and client application are implemented on a PC, while the landing algorithm, which is the design under test, on the Arty Z7 board.}
    \label{fig:HiL scheme}
\end{figure}

The system we propose consists of two main components.
The first one is a PC class computer equipped with a high-end GPU.
It is used to run the AirSim simulator server \cite{airsim} and the client programme, which also allows serial communication via the USB port.
We decide to use that simulator because of its high-quality graphics, generated using Unreal Engine 4.
This feature is crucial for the vision-based algorithm described in \ref{subsec:algorithm}.
The second component is the device under test (DUT), which is the Arty Z7 heterogeneous SoC FPGA computing platform.
It is built around the Zynq-7000 XC7Z020-1CLG400C chip from AMD Xilinx and HDMI video input and output ports.
The control algorithm implemented on the board enables the automatic landing of the drone on a detected landing site.
In addition to the elements mentioned above, we also use two LCD monitors. 
One is used to visualise the generated 3D environment, while the other is used to display the result of the landing site detection algorithm.
The schematic of the entire system is shown in Figure \ref{fig:HiL scheme}.
We describe its individual components in more detail below.


\subsection{Landing strip detection \& control algorithm}
\label{subsec:algorithm}

\begin{figure}[!t]
    \begin{minipage}{0.45\textwidth}
     \centering
     \begin{subfigure}[b]{0.71\textwidth}
         \centering
         \includegraphics[width=\textwidth]{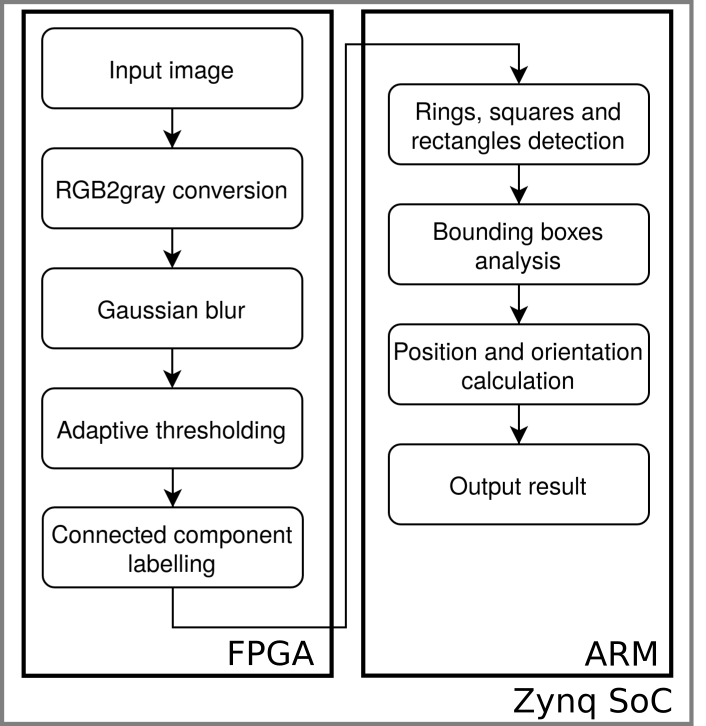}
         \caption{}
         \label{fig:detection scheme}
     \end{subfigure}
     \hfill
     \begin{subfigure}[b]{0.27\textwidth}
         \centering
         \includegraphics[width=\textwidth]{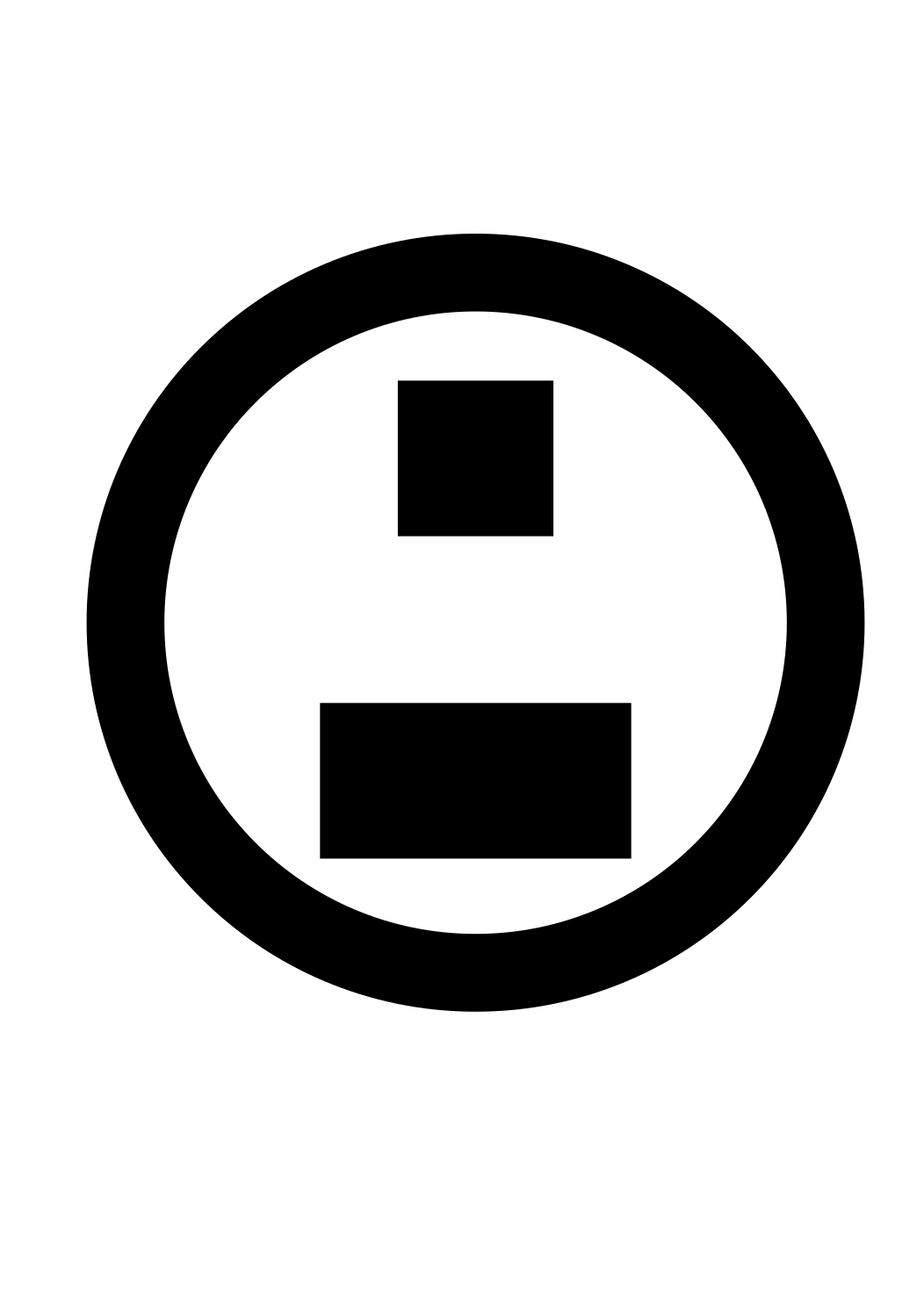}
         \caption{}
         \label{fig:landing strip}
     \end{subfigure}
        \caption{Scheme of the landing strip detection algorithm implemented in the Zynq SoC device (a) and the used marker (b).}
        \label{fig:vision alg}
    \end{minipage}
\end{figure}

The control algorithm that we used as case study is based on one of our previous works \cite{blachut2020}.
In it, we prepared a vision-based landing site detection system (see Fig. \ref{fig:detection scheme}), which we implemented on the Arty Z7 platform.
As part of this work, we made some modifications due to the specificity of the image generated by the simulator used.
We also changed the landing strip used -- its current version is shown in Fig. \ref{fig:landing strip}.
Compared to the paper \cite{blachut2020}, it differs in the absence of an inner ring (it was placed exactly in the centre of the marker).
This is due to a sometimes observed malfunction -- incorrect detection as the upper square.


The marker detection algorithm consists of the following steps. 
First, the image is converted to greyscale and subjected to low-pass filtering with a Gauss kernel.
Next, adaptive thresholding is implemented, which performs better under potentially nonuniform lighting conditions.
In the first step, the image is divided into non-overlapping squares in which the average brightness is calculated.
Based on this, a binarisation threshold is calculated.
In the second step, the threshold for each pixel is calculated on the basis of bilinear interpolation -- see the paper \cite{blachut2020} for details.
The next step is connected components labelling -- each object is described by the area, centroid, and the coordinates of the bounding box.
The elements mentioned so far are implemented in the reprogrammable part of the SoC FPGA chip and work in real time, i.e. they process the input video stream $1280 \times 720$ @ 60fps.
The object data are then sent to the processor system, where they are analysed.
On the basis of these data, the elements ring, square, and rectangle are detected.
This allows the position and orientation of the drone in relation to the marker to be determined.



The described vision algorithm requires information about the sizes of the geometric figures that are searched, in order to verify the correctness of the detection.
These values depend on the height at which the drone is located, as well as on the parameters of the camera and lens used.
The altitude measurement is carried out by a laser distance sensor located on the drone.



Please note, that the described vision algorithm was prepared and implemented in our previous work \cite{blachut2020}.
We use it as a~case study for our hardware-in-the-loop system.
To do so, we extend it with a simple control algorithm.
The vision algorithm returns three values: the two $(x, y)$ coordinates of the landing strip in the image, and its orientation $\theta$.
We use these to determine the control signals. 
In our case, these are the linear airspeeds $(v_x, v_y, v_z)$ and the rotation speed around the vertical axis of \textit{yaw} ($\omega_{yaw}$) in the local NED (North-East-Down) reference system associated with the drone. 
We use common P controllers, with gains chosen heuristically, to determine these:
\begin{align}
    v_x &= K_1 * \Delta x \label{eq:PIDx}\\
    v_y &= K_1 * \Delta y \label{eq:PIDy}\\
    v_z &= K_2 * h \label{eq:PIDz}\\
    \omega_{yaw} &= K_3 * \Delta \theta \label{eq:PIDyaw} 
\end{align}
where: $K_i$ -- gain values of the controllers; $\Delta (*)$ -- difference between the current position (linear/angular) and the target position; $h$ -- height at which the drone is currently located.


The landing procedure is divided into two stages.
In the first one, the drone has to move directly over the landing site and align itself according to its orientation.
At that time, the regulator \eqref{eq:PIDz} is turned off and $v_z = 0$.
This ensures that the vehicle maintains a constant altitude and does not lose the landing site from the camera field of view.
In the second stage, the drone performs a direct landing manoeuvre.
The previously deactivated regulator, \eqref{eq:PIDz}, is responsible for the descent, while \eqref{eq:PIDx} and \eqref{eq:PIDy} control the position of the vehicle so that it lands on the marker.
Just above the ground (at the set height $H_L$), a~command is sent that activates the vertical landing procedure in the lower level controller, which is responsible for directly controlling the vehicle's engines.
The entire control algorithm is implemented in the ARM Cortex-A9 processor, available inside the Arty Z7 SoC.


It should be noted that the development of an optimal (in any sense) control algorithm is not the main objective of this work.
It is only an intermediate step that enables the automation of drone movement.
This allows for meaningful experiments regarding HiL simulations for control algorithms implemented on the heterogeneous computing platforms.


We also implement a simple message detector on the processor for messages received over the serial link (details are described in Section \ref{subsec:communication}).
Its operation is limited to quickly scanning the input buffer for predefined strings.
This allows to provide the vision algorithm with the needed additional information (e.g. the current altitude of the drone) without significantly affecting the normal duty cycle.


\subsection{Simulator API}

\begin{figure}
    \centerline{\includegraphics[width=0.47\textwidth]{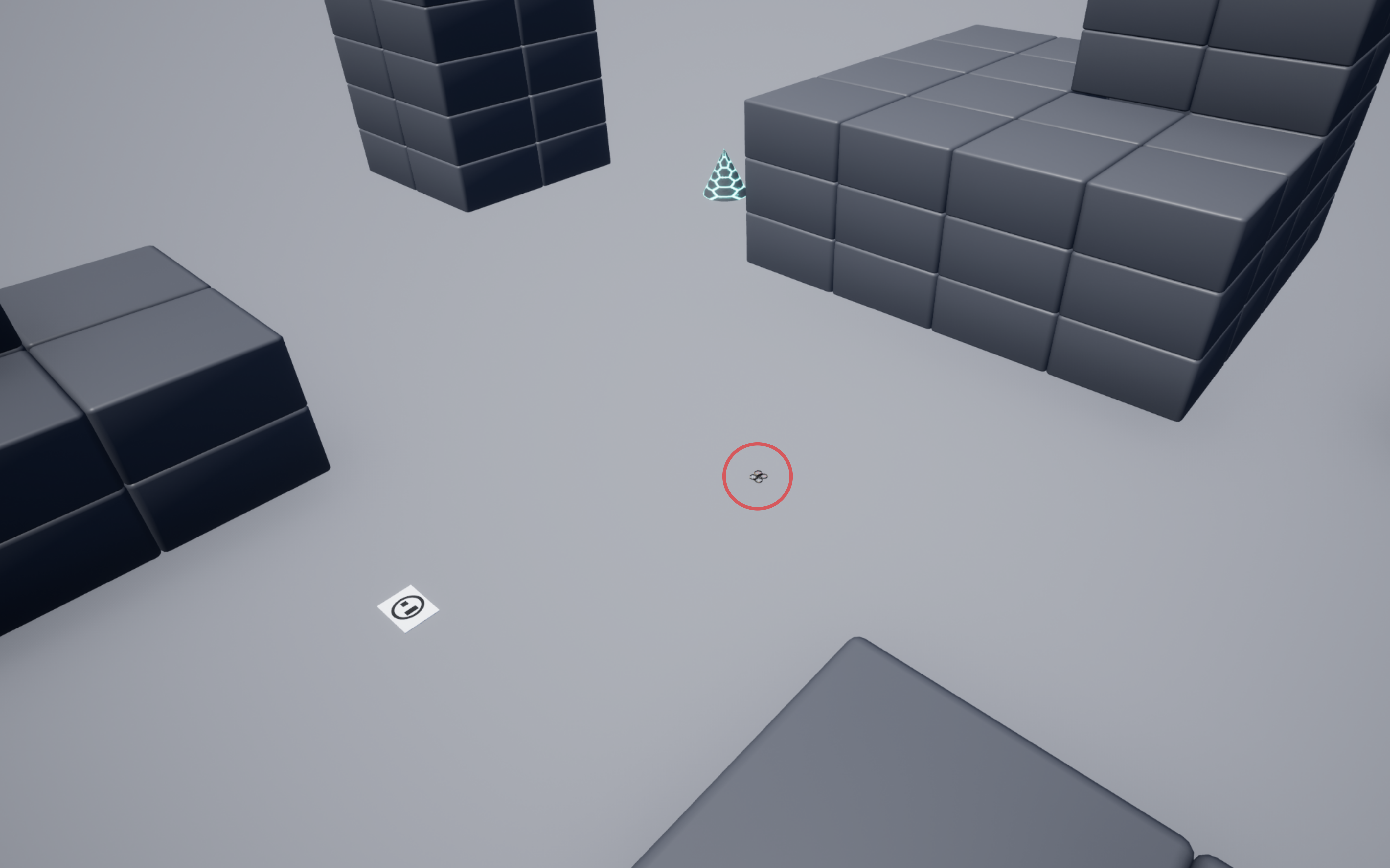}}
    \caption{Sample image of the \textit{Blocks} environment in the AirSim simulator. The drone model visible inside the red circle.}
    \label{fig:blocks env}
\end{figure}

As we have already mentioned, in our work we use the AirSim simulator, dedicated, among others, to unmanned flying vehicles.
It allows not only the generation of a virtual 3D environment but also simulates the vehicle dynamics and many typical sensors.
For this work, we use the default test environment \textit{Blocks}.
We only add to it the landing pad marker described earlier.
A sample image of our environment is shown in Figure \ref{fig:blocks env}.


The drone we are using is the basic \textit{SimpleFlight} model.
However, we modify it by adding a camera with HD resolution (1280 $\times$ 720), pointing perpendicularly downward.
The video stream received from it is the input to the vision algorithm described in Section \ref{subsec:algorithm}.
In addition to this, we add a distance sensor to the drone.
It is positioned and pointed exactly like the aforementioned camera and provides information about the current altitude at which the vehicle is located.


The environment, together with the drone model, is run as an AirSim server.
In addition, we have prepared a client application in Python language, using the dedicated Image API \cite{ImageApi} and the OpenCV library \cite{opencv}.
It mainly allows us to retrieve basic data from the simulator, namely the video stream and the current altitude information.
Moreover, it allows to send and receive data transmitted via the USB port of the PC.


\subsection{Communication}
\label{subsec:communication}

Data exchange between the elements of our system takes place through two channels.
The video stream is transmitted via HDMI.
In this way, we replace the physical camera at the input of the video algorithm with the image generated in the AirSim simulator.
Other data is exchanged through the serial protocol over a USB connection.
Among them, we should point out first of all: the current altitude of the drone (PC -> ArtyZ7) and the control signals \eqref{eq:PIDx}--\eqref{eq:PIDyaw} (ArtyZ7 -> PC).


To distinguish between the transmitted information, we develop a simple protocol.
The messages sent by the PC have the form \texttt{C(c/d)dd}, where \texttt{C} denotes the type of data, and the three decimal digits \texttt{d} encode the specified value.
Sometimes the first of them more precisely specifies the type of data being transmitted -- the case of \texttt{c}.
For example, transmitting the current altitude of $9.87$ [m] requires two messages: \texttt{Am09}, which transmits the integer part (meters), and \texttt{Ac89}, which transmits the fractional part (centimeters).


Communication in the opposite direction (from ArtyZ7 to PC) is similar.
This time, the messages are in the form \texttt{c:\textit{data}\textbackslash n}, where \texttt{c} denotes the data type, \texttt{\textit{data}} is its specified value, and \texttt{\textbackslash n} is the end symbol.
For example, transmission of control signals for speed ($v_x = 3.456, v_y = 7.892, v_z = 1.936$) amounts to the following message: \texttt{v:3.456,7.892,1.936\textbackslash n}.


The noticeable difference between the forms of messages sent in different directions is due to the far greater computational capabilities of the PC relative to the SoC device.
For this reason, we use a very simple structure of messages transmitted to the DUT, so that the overhead associated with their decoding does not interfere with the normal mode of operation of the embedded platform.


\section{Results \& discussion}
\label{sec:evaluation}

\begin{figure}[!t]
    \begin{minipage}{0.45\textwidth}
     \centering
     \begin{subfigure}[b]{0.49\textwidth}
         \centering
         \includegraphics[width=\textwidth]{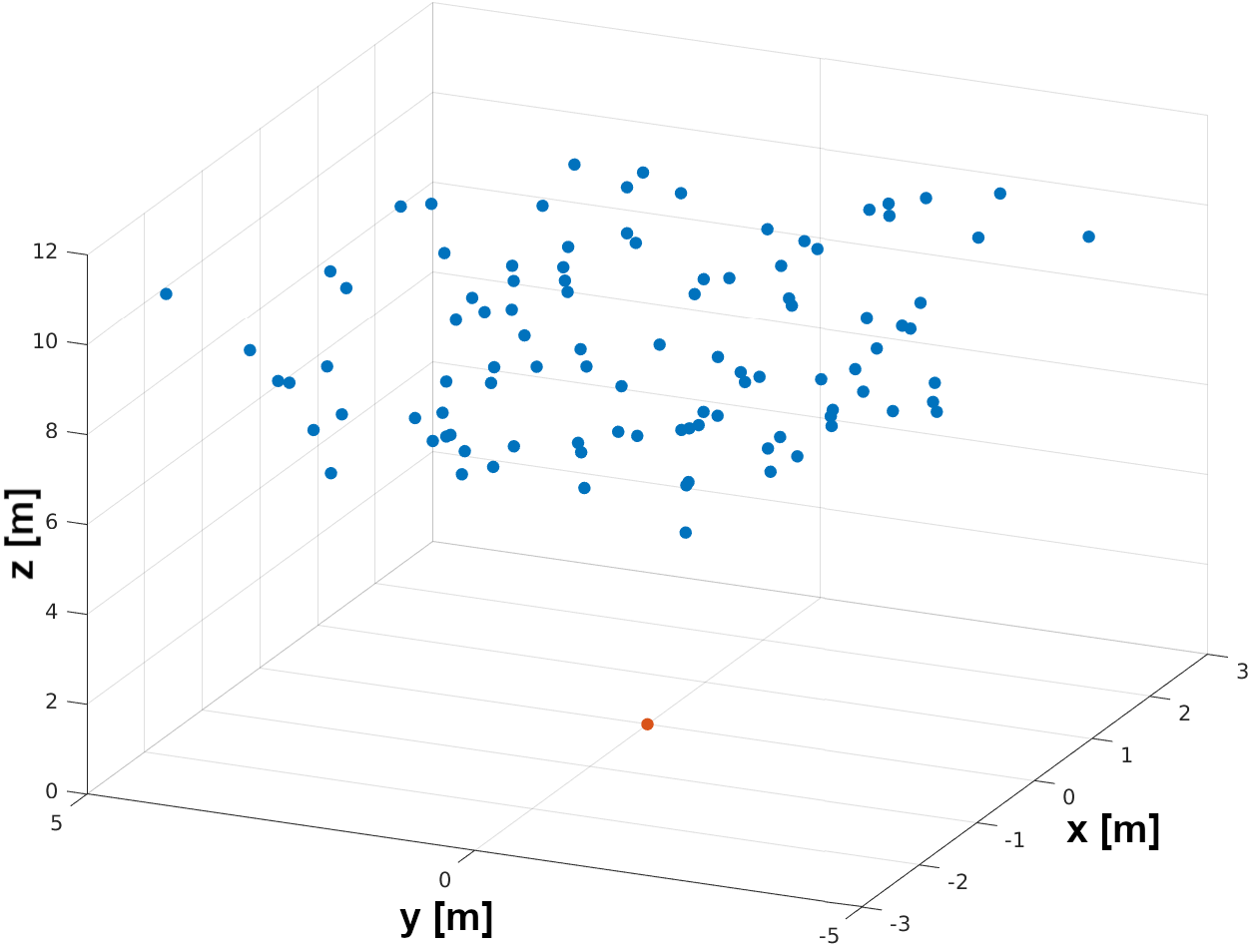}
         \caption{}
         \label{fig:start 3D}
     \end{subfigure}
     \hfill
     \begin{subfigure}[b]{0.49\textwidth}
         \centering
         \includegraphics[width=\textwidth]{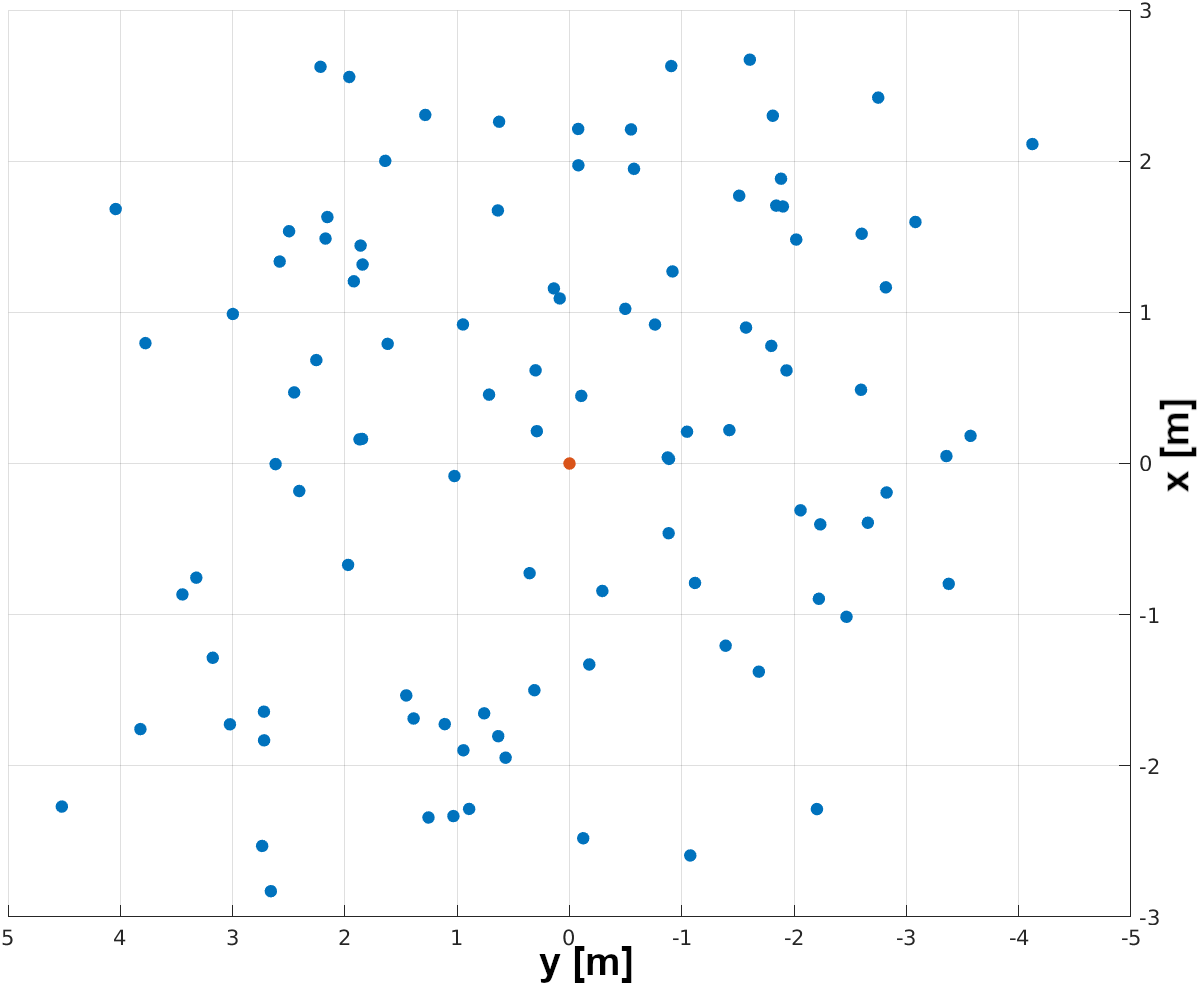}
         \caption{}
         \label{fig:start 2D}
     \end{subfigure}
        \caption{Visualisation of the 3D distribution of the take-off points (blue) (a) and their projection onto the XY plane of the landing field (b). The red colour indicates the centre of the landing site (target point).}
        \label{fig:start points}
    \end{minipage}
\end{figure}

We use the system we have developed to perform drone landing tests from randomly selected start positions and initial orientations.
However, due to the nature of the algorithm, we introduce some limitations.
First, we assume an initial altitude range of 5 [m] to 10 [m].
This ensures the possibility of correct detection of the landing marker (in practise, this range is strictly determined by the parameters of the camera used and its lens).
The remaining spatial coordinates depend on the selected altitude.
This is done so that the marker remains in the field of view of the camera and its detection is possible.
We do not impose any restrictions on the initial orientation of the drone (we allow all possible \textit{yaw} angles).
We repeat the procedure 100 times and obtain the distribution of start points shown in Figure \ref{fig:start points}.


\begin{figure}[!t]
    \centerline{\includegraphics[width=0.47\textwidth]{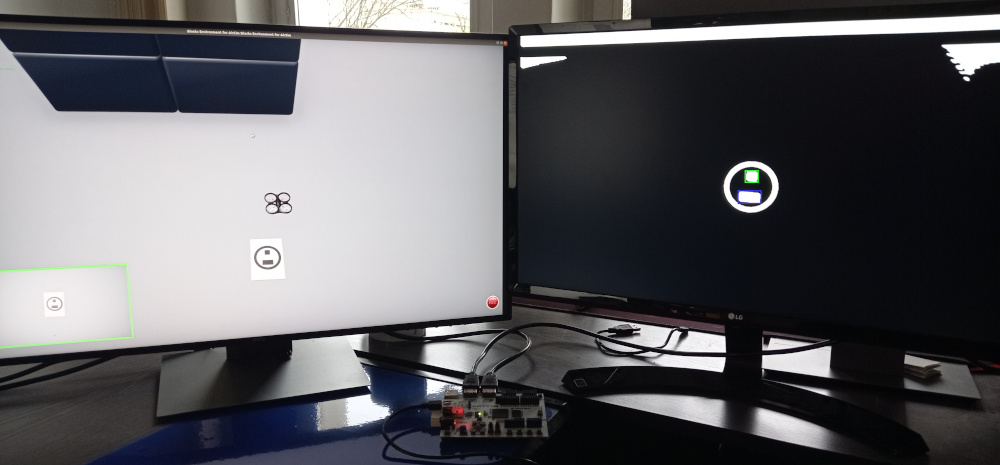}}
    \caption{Our system during the  HiL simulation. The left monitor shows the current state of the environment, while the right monitor shows the output of the landing site detection algorithm with the detected geometric figures. In the middle, below the monitors -- the Arty Z7 SoC FPGA board}
    \label{fig:hil working}
\end{figure}

At each point selected as described above, we run the control algorithm with the special command \texttt{T111} (see description in Section \ref{subsec:communication}).
This is equivalent to the situation where the landing procedure has been started after bringing the drone to the vicinity of the landing site, e.g. using GPS navigation.
Then we wait until the manoeuvre is completed.
In doing so, we consider two main possibilities.
In the first case, a malfunction of the simulation causes the drone to run away from the landing area.
In the end, it disappears from the field of view of the camera, and further action does not make sense.
In this case, we abort the iteration and treat the result as a failure.
The second possibility to complete the manoeuvre is a successful landing on the ground.
Then we measure the final position and orientation relative to the marker, which we treat as an error.
In addition, we measure the time from the start of the landing procedure to its completion.
We collect the results obtained in Table \ref{tab:ending types}, Table \ref{tab:land precision} and Table \ref{tab:landing time}.
Furthermore, in Figure \ref{fig:hil working} we present our system during one of the test cases.


\begin{table}[!t]
\caption{Landing procedure completion achieved.}
\begin{center}
\begin{tabular}{|l|c|c|}
\hline
Ending type & Quantity & Percentage [\%] \\
\hline
Success & 100 & 100\\
\hline
Failure & 0 & 0 \\
\hline
\end{tabular}
\label{tab:ending types}
\end{center}
\end{table}

\begin{table}[!t]
\caption{The average landing accuracy (mean square error) obtained. For comparison: the inner radius of the ring, which is part of the marker \ref{fig:landing strip}, is 0.7 m.}
\begin{center}
\begin{tabular}{|l|c|}
\hline
 & MSE \\
\hline
x [m] & 0.3375 \\
\hline
y [m] & 0.8672 \\
\hline
$\theta$ [rad] & 0.7505 \\
\hline
\end{tabular}
\label{tab:land precision}
\end{center}
\end{table}

\begin{table}[!t]
\caption{Landing procedure completion time.}
\begin{center}
\begin{tabular}{|c|c|c|c|}
\hline
Average [s] & Median [s] & Maximum [s] & Minimum [s]\\
\hline
43.02 & 43.83 & 59.38 & 25.89\\
\hline
\end{tabular}
\label{tab:landing time}
\end{center}
\end{table}

The results obtained mainly demonstrate the correctness of the proposed system for HiL simulation of control algorithms implemented on heterogeneous computing platforms.
In all the test scenarios considered, the drone remained in the area where the landing marker is in the field of view of the camera.
This demonstrates the correct data exchange between the simulator (PC) and the control unit (Arty Z7).
Moreover, the average landing precision and achieved time statistics indicates a~sufficient level of the synchronisation between PC and Arty Z7.
Otherwise, significant oscillations (potentially of unsteady period) would be observable, resulting from the lack of adequate updating of the transmitted information (e.g. control).
Such a precise and repeatable landing would not then be possible.


The only noticeable problem is the relatively large final orientation error.
We can point to at least two explanations for this.
First, we should note the rather simple structure of the control algorithm (as it was only used to verify the performance of the simulation).
The moment of transition from the phase of flight over the marker to the descent stage plays a major role in it.
The resulting inaccuracies in orientation are transferred directly to the final result.
The second potential cause is the lack of proper filtering of current altitude measurement.
Based on its instantaneous value (even after slight averaging) may lead, in the situation of measurement disturbance, to premature detection of the $H_L$ level and the initiation of a direct landing procedure before the orientation is properly determined.
The correct cause is probably a combination of both sources indicated.
However, it should be noted that from the perspective of the expected results of the landing procedure, the final location of the drone is much more important than its orientation.
Furthermore, the development of an optimal control algorithm was not the aim of this work - simple control algorithm was used only for the verification of the performance of the simulation.


It is also worth noting the adequate graphical performance of the simulator.
This is because the video stream was generated at a sufficiently high image rate, which enabled the correct operation of the landing site detection algorithm adapted to process as much as 60 FPS.
On the other hand, during development of the system, we used a software-in-the-loop simulation.
In this case, the landing strip detection application in Python allowed only for several frames per second.
This is just another example of real-time performance of FPGA based video stream processing.


\section{Conclusions \& future work}

In our work, we have designed and successfully implemented a HiL simulation system for UAV control algorithms implemented on heterogeneous SoC FPGA platforms.
Our proposal uses only a PC and a USB serial link, making it relatively inexpensive and applicable in almost any similar situation.
The results obtained, in turn, prove the correct operation of such a system.
This is due, among other things, to the simple and effective protocol for data exchange via the serial link.
Due to its design, it does not overload the tested algorithm.
However, it is worth noting that even during the target operation, the flying platform used will have to communicate with a lower-level controller.
In addition, the use of a heterogeneous computing platform made it possible to significantly speed up the operation of the vision algorithm while reducing the power requirements.


We plan to use this HiL simulation system in the course of our work on various UAV control algorithms.
Among the potential areas of application, in addition to the discussed autonomous landing problem, we can point to, among others, autonomous drone racing, exploring unknown areas (SLAM -- Simultaneous Localisation and Mapping) or automatic terrain inspection.
Moving from simulation to reality will also enable the tool itself to be further improved and better adapted to real working conditions.
The use of a SoC FPGA should allow us to implement computationally complex algorithms, even deep neural networks (DNNs) in real time.


\bibliographystyle{IEEEtran}
\bibliography{sk_spa_2022.bib}

\end{document}